\RequirePackage{fix-cm}
\documentclass[twocolumn]{svjour3}          % twocolumn
\smartqed  % flush right qed marks, e.g. at end of proof
\usepackage{graphicx}
\journalname{CGI2026} % The correct name will be entered by the editor

\usepackage{tabularray}
\usepackage{subfig}
\usepackage{color}
\usepackage{tabularx}
\usepackage{amsmath}
\usepackage{amssymb}
\usepackage{booktabs}
\usepackage{multirow}
\usepackage{makecell}

\usepackage{xspace}
\newcommand{\RADN}{RADN\xspace}
\newcommand{\HATSNet}{HATSNet\xspace}
\newcommand{\FBHR}{FBHR\xspace}
\newcommand{\TotalReLighting}{TR\xspace}
\newcommand{\SwitchLight}{SL\xspace}
\newcommand{\AFHR}{AFHR\xspace}

\newcommand{\Ours}{HAFMat\xspace}
\newcommand{\HumatDataset}{OpenHumanBRDF\xspace}
\newcommand{\HuMat}{HM\xspace}
\newcommand{\Dino}{DinoV2\xspace}

\newcommand{\FusionFullLow}{multi-layer adaptive feature fusion mechanism\xspace}
\newcommand{\FusionFull}{Multi-layer Adaptive Feature Fusion Mechanism\xspace}
\newcommand{\FusionFullShort}{MAFFM}

\begin{document}

\title{\Ours: Hybrid Priors Guided Adaptive Fusion for Single-Image Human Material Estimation}
\subtitle{}
\author{Yu Jiang, Jiahao Xia, Jiongming Qin, Jianchi Sun, Chunxia Xiao$^{*}$}
\institute{
	Yu Jiang, jiangyu1181@whu.edu.cn\\
	Jiahao Xia, jiahao.xia-1@uts.edu.au\\
	Jiongming Qin, 2022102110003@whu.edu.cn\\
	Jianchi Sun, sunjc0306@whu.edu.cn\\
	Chunxia Xiao, cxxiao@whu.edu.cn\\
	*Corresponding author: Chunxia Xiao. \\
	Yu Jiang, Jiongming Qin, Jianchi Sun, and Chunxia Xiao. School of Computer Science, Wuhan University, Wuhan 430072, Hubei, China. \\
	Jiahao Xia. Faculty of Engineering and IT, University of Technology Sydney, Australia.
}
\date{2026}% The correct dates will be entered by the editor

\maketitle

\begin{abstract}
Physically based rendering (PBR) material estimation is a fundamental appearance decomposition task with broad applications in virtual content creation, relighting, and digital human rendering. However, estimating PBR materials from a single human image remains highly ill-posed, since illumination, geometry, and reflectance are heavily entangled in the observed appearance. 
To mitigate this ambiguity, we propose \Ours, a hybrid-prior-guided framework for single-image human material estimation. Our method introduces guidance maps that encode complementary cues, including appearance, body geometry, structure, and prior material predictions from pre-trained models. 
A key observation is that these guidance cues are heterogeneous. Some cues primarily provide texture-level constraints. Others convey higher-level semantic information.
To exploit this property, we design a \FusionFull, which adaptively fuses guidance features with decoder features at different stages. This design enables texture-dominant and semantic-dominant cues to guide material decoding at appropriate levels, leading to more accurate and physically plausible material estimation. 
Extensive experiments on both synthetic and real data demonstrate that our method achieves state-of-the-art performance in material estimation and downstream relighting.

\keywords{Physically Based Rendering \and Inverse rendering \and Human Relighting}
\end{abstract}

\begin{figure}[!t]
	\centering % 居中
	\begin{minipage}[b]{\linewidth} % 整体划分一个页面
		\centering % 居中
		\newcommand{\myvspace}{1.0 pt} % 定义行之间的空隙
		\newcommand{\widthOfFullPage}{1} % 1 / 列数
		\newcommand{\widthOfMiniPage}{0.99}
		\newcommand{\format}{png}
		\begin{picture}(0,0)
			%			\footnotesize
			\put(40,-28){\rotatebox{90}{\makebox(0,0)[c]{\scriptsize w/o}}}
			\put(40,-85){\rotatebox{90}{\makebox(0,0)[c]{\scriptsize w/}}}
			\put(6,3){\makebox(0,0)[c]{\tiny \HuMat}}
			\put(25,3){\makebox(0,0)[c]{\tiny \AFHR}}
		\end{picture}
		\includegraphics[width=\widthOfMiniPage\linewidth]{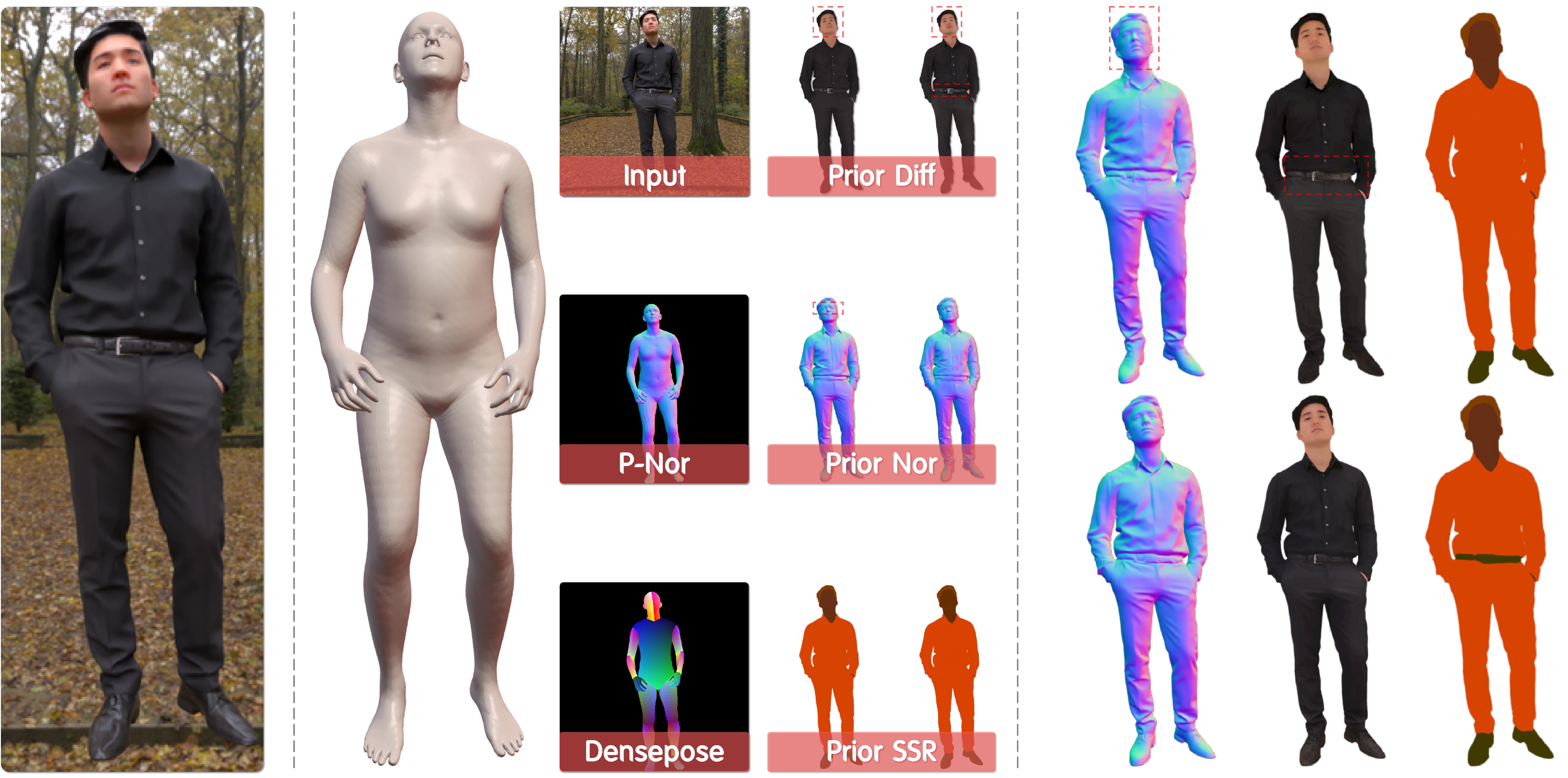}
		\begin{flushleft}
			\small
			\hspace{1.3em}Input
			\hspace{2.3em}SMPL
			\hspace{0.7em}Guidance Maps
			\hspace{1.6em}Nor
			\hspace{0.9em}Diff
			\hspace{0.9em}RSS
		\end{flushleft}
	\end{minipage}
\caption{
	\Ours. Our method aims to estimate PBR materials from a single image. The estimated material maps include ``Nor'' (normal), ``Diff'' (diffuse albedo), and ``RSS'' (subsurface scattering, specular albedo, and roughness). The ``Guidance Maps'' consist of the input image, geometry-related maps (``P-Nor'' and ``DensePose'') derived from the SMPL model reconstructed by PyMAF~\cite{pymaf2021,pymafx2023}, and prior material maps predicted by the pre-trained \HuMat~\cite{HumanMaterial} and \AFHR~\cite{AFHR_2025_EGSR}. ``w/" indicates that the estimated results are obtained by employing these guidance maps in the decoding process, while ``w/o" means they are not.
}
\label{img-teaser}
\end{figure}
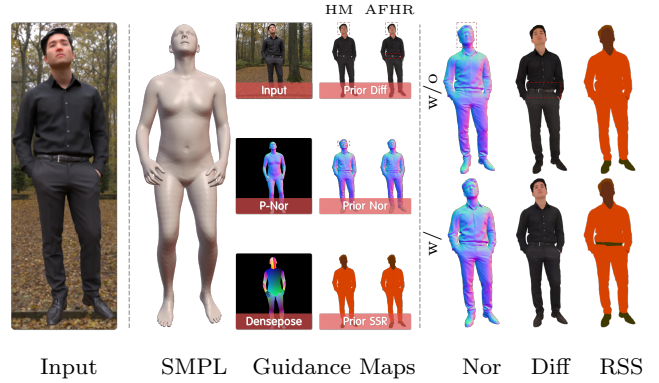 
\section{Introduction}
\label{sec:introduction}

Physically based rendering (PBR) material estimation is a fundamental appearance decomposition task with broad applications in relighting, digital humans, virtual content creation, and game production~\cite{SwitchLight_2024_CVPR_pbr_and_neural_render,MaXiaoHe_TVCG_2024}. Given one or more images, the goal is to recover physically meaningful material properties that enable realistic rendering under novel illumination. However, estimating PBR materials from a single human image remains highly ill-posed, since illumination, geometry, and reflectance are heavily entangled in the observed appearance~\cite{AFHR_2025_EGSR,RADN_2018_TOG_single_image_natural_svbrdf,SwitchLight_2024_CVPR_pbr_and_neural_render,HumanIR_MGI_2025_TVC,WRICNet_2022_TGRS,QinJiongMing_2025_SigAsia,SunJianChi_2026_AAAI,CaoTuo_2025_CVPR,WangYuSen_2025_MM,WangYuSen_2022_SigAisa}. The problem is further complicated in human scenes by the coexistence of diverse materials, such as fabric, leather, and skin, whose reflectance behaviors differ substantially.

Existing methods attempt to mitigate this ambiguity by constructing dedicated datasets~\cite{FBHR_2021_EGSR,TotalRelighting_2021_Sig_single_image_human_phong_neural_render,SwitchLight_2024_CVPR_pbr_and_neural_render,HumanMaterial,AFHR_2025_EGSR} or introducing prior knowledge from pre-trained models~\cite{HumanMaterial}. While these strategies have improved single-image human material estimation, the task remains challenging because the visual evidence available in a single image is still severely limited. In particular, reliable material prediction often requires complementary cues beyond RGB appearance alone, including scene illumination, body geometry, human structure, and prior material knowledge. Effectively incorporating such complementary information is therefore critical for reducing ambiguity and improving physical plausibility.

A key observation in our work is that these guidance cues are inherently heterogeneous. Some cues, such as geometry-derived normals and prior material maps, mainly provide local texture-level constraints and are most useful for preserving fine material details. In contrast, appearance cues and body-structure cues provide higher-level semantic information, such as illumination layout and material-category distributions. This heterogeneity makes naive feature fusion suboptimal: uniform fusion can either overemphasize local texture details at the expense of semantic context or overly favor semantics while weakening fine-grained material cues. Therefore, an effective material estimation framework should not only incorporate hybrid guidance, but also adaptively fuse different guidance cues at appropriate representational levels.

To this end, we propose \Ours, a hybrid-prior-guided framework for single-image human material estimation. As shown in Fig.~\ref{img-teaser}, our method introduces a set of guidance maps that encode complementary appearance, geometry, structure, and prior-material information. To exploit their heterogeneity, we further design a \FusionFullLow~(\FusionFullShort), which adaptively integrates guidance features with decoder features at different stages. In this way, texture-dominant and semantic-dominant cues can guide material decoding at the levels where they are most informative. Extensive experiments on both synthetic and real data demonstrate that our method achieves state-of-the-art performance in material estimation and downstream relighting.

To summarize, we make the following main technical contributions.
\begin{itemize}
	\item Present a hybrid guidance pipeline for single-image human material estimation that combines SMPL-derived geometric and structural cues with prior material predictions from pre-trained models.
	\item Introduce a set of guidance maps that encode geometric cues, structural constraints, and material priors, guiding material decoding toward physically plausible estimates.
	\item Design \FusionFullShort ~to identify guidance-map characteristics and adaptively fuse heterogeneous guidance features at appropriate levels, thereby improving material estimation accuracy.
\end{itemize}

\section{Related Work}
\label{related_work}
Inverse rendering is a fundamental topic in computer graphics and centers on two core tasks: PBR material estimation and relighting. The two are closely intertwined, as accurate material decomposition enables high-fidelity relighting, while relighting quality in turn depends on reliable appearance and material estimation. This connection is particularly important for human images, where illumination, geometry, and reflectance are strongly entangled.

\textbf{PBR Material Estimation} aims to recover spatially varying reflectance properties that conform to physical laws, enabling realistic rendering under arbitrary lighting~\cite{AvatarMe++_2022_TPAMI_Lattas,Relightify_2023_ICCV_DiffusinModel_SVBRDF_facial,MaXiaoHe_TVCG_2024}. Estimating PBR materials from a single image remains challenging due to the severely ill-posed nature of the problem~\cite{Li_2018_TOG_single_image_natural_svbrdfs,Ye_2018_CGF_single_image_plane_natural_svbrdf}. Early works focus on near-planar surfaces~\cite{RADN_2018_TOG_single_image_natural_svbrdf,DIR_2019_TOG_single_image_natural_svbrdf,MaterialGAN_2020_TOG_single_image_natural_svbrdf,HATSNet_2021_TOG_single_image_natural_svbrdf,LATNet_2022_TOG_single_image_natural_svbrdf}. For human scenarios, some works alleviate the ill-posed issue by using prior information~\cite{SwitchLight_2024_CVPR_pbr_and_neural_render} or constructing dedicated human material datasets~\cite{AFHR_2025_EGSR,HumanMaterial}. 
%
%Multi-image-based methods~\cite{NeILF_2022_ECCV_Multi_images_natural_svbrdf,Nvdiffrec_2022_CVPR_natural_multi_images_svbrdf,InvRender_2022_CVPR_Multi_images_natural_svbrdf,TensoIR_2023_CVPR_Multi_images_natural_svbrdf,Factored_NeuS_2025_CVPR_Multi_images_natural_svbrdf} leverage viewpoint variations to constrain material reconstruction. Combining data-driven BRDF priors~\cite{Nerfactor_2021_SigAsia_multi_images_natural_brdf}, reconstructing geometry and materials in different stages~\cite{NeRO_2023_Sig_Multi_images_natural_svbrdf}, and adopting more efficient representations~\cite{Nvdiffrec_2022_CVPR_natural_multi_images_svbrdf,TensoIR_2023_CVPR_Multi_images_natural_svbrdf} are all helpful for material reconstruction. However, most existing methods either focus on generic objects and surfaces or rely on multi-view observations. For human scenarios, single-image PBR material estimation remains highly ambiguous, and the complementary roles of different guidance cues have not been fully explored.

\begin{figure*}[!ht]
	\centering
	\includegraphics[width=\textwidth]{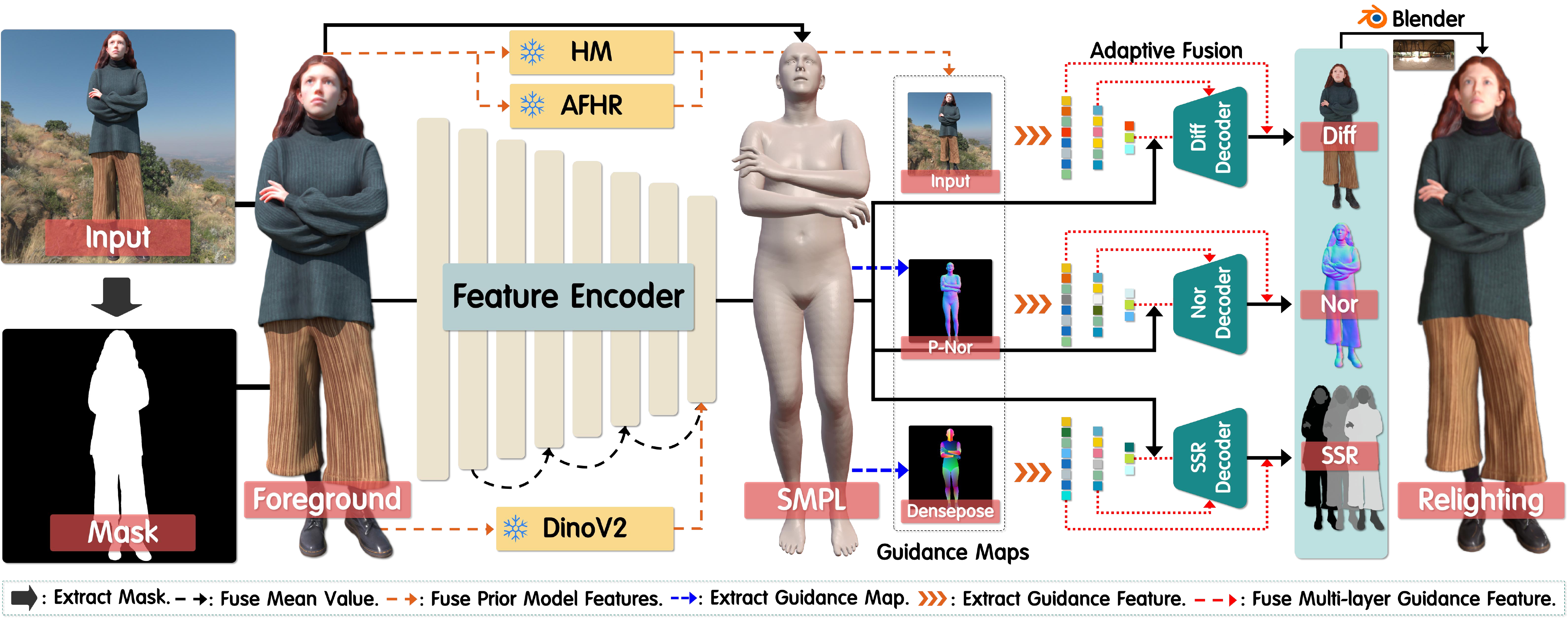}
	\caption{
	Method overview. 
	Our method, \Ours, estimates five PBR material maps from a single human image.
	First, we acquire the ``Mask" and ``Foreground" from the ``Input". 
	We design a ``Feature Encoder" to extract latent features of the ``Foreground", and extract  ``P-Nor" and ``Densepose" maps from the estimated SMPL model. Moreover, we fuse features from three pre-trained prior models (DinoV2, \HuMat and \AFHR).
	During the decoding process, we propose an effective \FusionFullLow to efficiently fuse the guidance features.
	Finally, we estimate the material maps and employ the Blender to achieve relighting under arbitrary illuminations.
}
\label{overview}
\end{figure*}
Recently, the generation of 3D assets with PBR materials from a single image has attracted increasing attention due to its practical value in applications such as game development and digital twins~\cite{MaterialMVP_ICCV_2025,Hunyuan3D25}. In such pipelines, single-image PBR material estimation serves as a core component for producing physically meaningful appearance, and existing generation models often rely on pre-trained estimation models~\cite{NeuralLightRig_2025_CVPR,RGBX_2024_SIG,MatFusion_2023_Siggraph,ControlMat_2024_TOG} to provide material priors. However, current 3D asset generation models still require large-scale data and computational resources. For human scenarios, the generated results are often inconsistent with human identity, and complex textures are prone to distortion. These limitations further highlight the need for explicit and reliable human PBR material estimation from a single image.

\textbf{Relighting} aims to synthesize images of humans under novel lighting~\cite{Relighting_humans_2018_SIGGRAPH,Single_image_portrait_relighting_2019_SIGGRAPH,Face_Relighting_2020_CVPR,Relight_Full_Body_2021_CGF,Face_Relighting_2022_CVPR,Full_Body_Relighting_2022_ECCV,Face_Relighting_2022_TPAMI,Photorealistic_2022_CVPR_single_image_human_neural_render}. For single-image human relighting, empirical models~\cite{FBHR_2021_EGSR}, neural renderers~\cite{TotalRelighting_2021_Sig_single_image_human_phong_neural_render}, and combined physical and neural shaders~\cite{AFHR_2025_EGSR} are widely adopted. Diffusion models possess strong generative capabilities~\cite{Sohl_2015_ICML_diffusionModel_basic_theory,DDPM_2020_DiffusionModels,DDIM_2020_DiffusionModels,CLIP_2021_TCML_BigDataModel,StableDiffusion_2022_DiffusionModels,Jain_2022_CVPR_DiffusionModel_3DGeneration,Magic3d_2023_CVPR_DiffusionModel_3DGeneration} and have been proven to achieve high-quality illumination harmonization~\cite{ICLight}, but they still struggle to provide physically accurate and controllable lighting manipulation. Moreover, extensive work has focused on achieving relighting based on multi-view data or videos~\cite{Relighting4D_2022_ECCV_video_human_svbrdf,Relightable_2023_ICCV_video_human_svbrdf,Relit_NeuLF_2023_ACMMM_Multi_images_facial_svbrdf}. While these methods have advanced relighting quality under various settings, photorealistic and controllable relighting for a single human image remains challenging. This also indicates that accurate estimation of physically meaningful human materials is still a crucial prerequisite for robust downstream relighting.

Human material estimation faces unique challenges due to the diversity of body materials (e.g., fabric, leather, and skin) and the complex reflectance behavior of skin in particular. Existing works suggest that priors such as subsurface scattering, human geometry, body structure, and prior material predictions can help reduce the ambiguity of single-image estimation. However, these cues are inherently heterogeneous: some mainly provide fine-grained texture or reflectance information, while others offer higher-level geometric or semantic guidance. Existing methods rarely distinguish these roles explicitly during material decoding. Our work is motivated by this observation and aims to better exploit such complementary cues for single-image human PBR material estimation.

\begin{figure*}[!ht]
	\centering % 居中
	\begin{minipage}[b]{\linewidth} % 整体划分一个页面
		\centering % 居中
		\newcommand{\myvspace}{1.0 pt} % 定义行之间的空隙
		\newcommand{\widthOfFullPage}{1} % 1 / 列数
		\newcommand{\widthOfMiniPage}{0.99}
		\includegraphics[width=\widthOfMiniPage\linewidth]{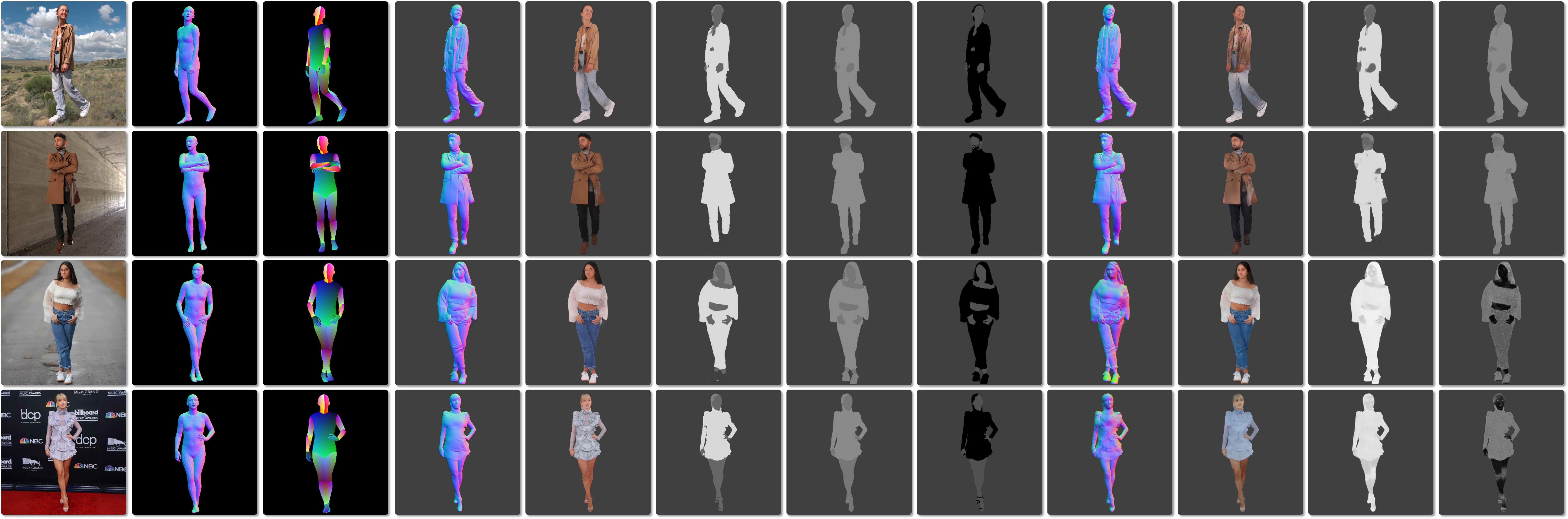}
		\begin{flushleft}
			\small
			\vspace{-0.5em}
			\hspace{1.3em}Input
			\hspace{1.8em}P-Nor
			\hspace{1.2em}D-Pose
			\hspace{1.8em}H-N
			\hspace{2.2em}H-D
			\hspace{2.3em}H-R
			\hspace{2.5em}H-S
			\hspace{2.2em}H-SSS
			\hspace{1.8em}A-N
			\hspace{2.3em}A-D
			\hspace{2.2em}A-R
			\hspace{2.5em}A-S
		\end{flushleft}
	\end{minipage}
	\caption{
		Samples of guidance maps. 
		``P-Nor" and ``D-Pose" refer to the normal and densepose maps extracted from the estimated SMPL model.
		``H-N", ``H-D", ``H-R", ``H-S" and ``H-SSS" refer to the normal, diffuse albedo, roughness, specular albedo, and subsurface scattering maps extracted from \HuMat~\cite{HumanMaterial}.
		``A-N", ``A-D", ``A-R", and ``A-S" refer to the normal, diffuse albedo, roughness, and specular albedo maps extracted from \AFHR~\cite{AFHR_2025_EGSR}.
		The above two samples are from \HumatDataset dataset, and the following two from real data.
	}
	\label{img-guidance_maps}
\end{figure*}
\section{Methodology}
\label{sec:method}
We propose \Ours, a hybrid-prior-guided framework for single-image human PBR material estimation. Given a single human image \(I \in \mathbb{R}^{H \times W \times 3}\), the goal is to predict five PBR material maps by learning a mapping function
\begin{equation}
	\mathcal{F}: I \mapsto (N, D, R, S, SSS),
\end{equation}
where \(N\), \(D\), \(R\), \(S\), and \(SSS\) denote the normal, diffuse albedo, roughness, specular albedo, and subsurface scattering maps, respectively. Specifically, \(N \in \mathbb{S}^{H \times W \times 3}\), \(D \in [0, 1]^{H \times W \times 3}\), and \(R\), \(S\), \(SSS \in [0, 1]^{H \times W \times 1}\). To reduce the ambiguity of single-image estimation, we introduce a set of guidance maps that provide complementary cues on appearance (including environmental lighting), geometry, body structure, and prior material properties. Since these cues are heterogeneous and contribute differently to material decoding, we further design \FusionFullLow~(\FusionFullShort) to adaptively fuse guidance and decoder features at different stages.

Although the \HumatDataset dataset~\cite{HumanMaterial} provides supervision for these PBR material maps, estimating all five maps from a single image remains highly ill-posed. To improve the physical plausibility of the estimated materials, we propose \Ours. Our framework contains an image feature encoder, a guidance encoder, and three material decoders for predicting the five PBR maps. In addition to the input image, we construct a set of guidance maps from SMPL reconstruction and pre-trained material estimation models. These guidance maps provide complementary geometry, structure, and prior-material cues. During decoding, their features are injected into the material decoders through \FusionFullShort, enabling different types of guidance to influence material prediction at appropriate representational levels.

As illustrated in Fig.~\ref{overview}, we first extract the ``Mask" and ``Foreground" from the input image, and design a ``Feature Encoder" to extract latent features from the ``Input". This encoder employs eight down-sampling blocks, with channel dimensions of 64, 128, 128, 128, 512, 512, 1024, and 1024. The large-scale pre-trained model \Dino~\cite{Dino_2023} has demonstrated strong feature extraction capability in visual representation learning. Therefore, we use pre-trained \Dino features extracted from the ``Foreground" and fuse them with the features from the ``Feature Encoder" to enhance appearance representation.
To constrain the decoded materials to a more reasonable range, we introduce guidance maps and guide the decoding process via adaptive feature fusion.
After decoding, we obtain five material maps, which can be used in rendering engines such as Blender for downstream applications including relighting and material editing.

\textbf{Guidance Maps.}
The visual evidence contained in a single human image is limited and highly entangled. Decoupling such limited information into five physically meaningful material maps is therefore highly ambiguous. To reduce this ambiguity, we construct guidance maps that provide complementary constraints for material decoding.

The five target material maps can be roughly grouped according to the type of information they require. Diffuse albedo is strongly related to appearance and shading. Normal estimation is closely related to geometry. Roughness, specular albedo, and subsurface scattering are more closely related to material category and body-part semantics. Accordingly, material estimation requires not only foreground appearance, but also environmental lighting context, geometric cues, and structural information. The ``Input" naturally provides appearance and lighting cues. Meanwhile, the geometry-aware normal map (``P-Nor") and the body-structure map (``Densepose") derived from SMPL can provide geometric and structural constraints, respectively.

Moreover, previous works~\cite{SwitchLight_2024_CVPR_pbr_and_neural_render,HumanMaterial,AFHR_2025_EGSR} have shown that competitive material maps can be estimated from a single image by training on dedicated human PBR material datasets. A direct strategy to further improve estimation accuracy is therefore to use such predictions as prior information to guide material decoding. We denote the hybrid guidance maps as
\begin{equation}
	\mathcal{G}=\{I, G_{\text{pnor}}, G_{\text{dp}}, \mathcal{P}_{HM}, \mathcal{P}_{AFHR}\},
\end{equation}
where \(I\) is the input image, \(G_{\text{pnor}}\) and \(G_{\text{dp}}\) denote the SMPL-derived normal and Densepose maps, respectively, and
\begin{equation}
	\begin{aligned}
		&\mathcal{P}_{HM}   = \{N^{hm}, D^{hm}, R^{hm}, S^{hm}, SSS^{hm}\}, \\
		&\mathcal{P}_{AFHR} = \{N^{af}, D^{af}, R^{af}, S^{af}\}.
	\end{aligned}
\end{equation}
Thus, as shown in Fig.~\ref{img-guidance_maps}, our guidance maps comprise 12 components in total.

\begin{figure}[!h]
	\centering % 居中
	\begin{minipage}[b]{\linewidth} % 整体划分一个页面
		\newcommand{\myvspace}{1.0 pt} % 定义行之间的空隙
		\newcommand{\widthOfFullPage}{1} % 1 / 列数
		\newcommand{\widthOfMiniPage}{1}
		\newcommand{\format}{png}
		\includegraphics[width=\widthOfMiniPage\linewidth]{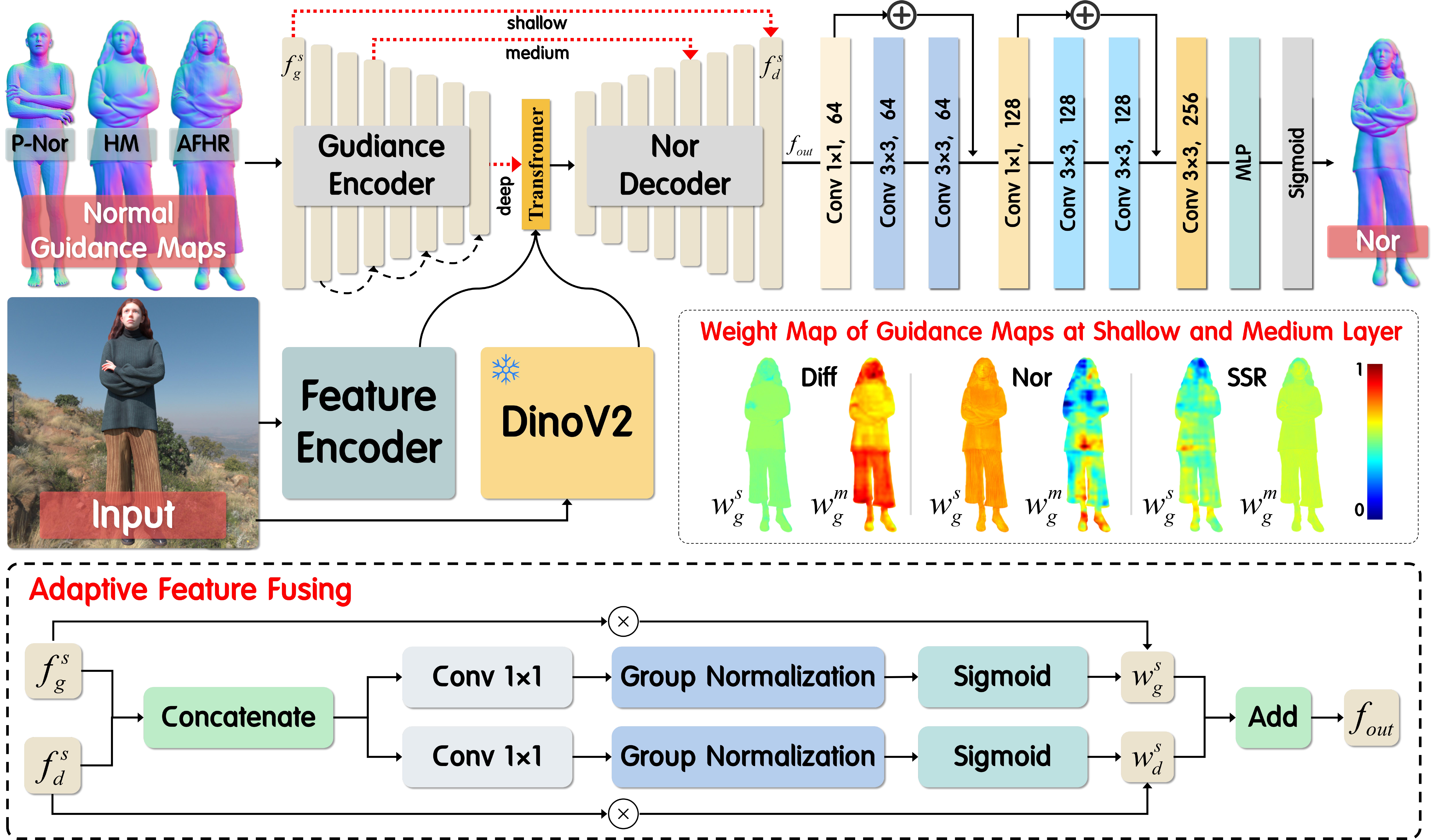}
	\end{minipage}
	\caption{
		The guided decoding process of the guidance map (P-Nor).
		$f_{g}^{s}$ and $f_{d}^{s}$ mean the shallow layer feature of Guidance Encoder and Nor Decoder, respectively.
		$w_{g}^{s}$ and $w_{d}^{s}$ mean the shallow layer weight of Guidance Encoder and Nor Decoder, respectively.
		$w_{g}^{m}$ means the medium layer weight of Guidance Encoder.
		$f_{out}$ means the feature after adaptive fusion.
		The decoding processes (Diff and SSR Decoder) guided by other guidance maps are similar to those guided by the ``Normal Guidance Maps".
	}
	\label{img-guidance_model}
\end{figure}

These hybrid guidance maps are complementary rather than redundant. As shown in Fig.~\ref{img-guidance_model}, ``P-Nor" provides more accurate facial geometry cues, \HuMat produces more globally reasonable material layouts, and \AFHR captures richer local details, such as fingers. Fully exploiting these complementary priors improves the accuracy and robustness of material estimation. We use both \HuMat and \AFHR because they are trained on different data distributions: one is based on real scanned human data, while the other is based on synthetic CG data. Their predictions therefore provide complementary prior knowledge for material decoding.

\textbf{Guided Material Decoding.}
As discussed above, the guidance maps are used to guide the decoding process of the three material decoders. These guidance maps differ substantially in both representational level and semantic content. Specifically, 
(1) the input image contains both shallow texture-level information and deeper semantic information;
(2) ``P-Nor" and prior material maps mainly provide pixel-level texture and reflectance constraints;
(3) ``Densepose" mainly provides body-structure information as a higher-level semantic cue.
We encode the hybrid guidance maps into multi-scale guidance features:
\begin{equation}
	\{f_g^{s}\}_{s=1}^{S} = \mathcal{E}_g(\mathcal{G}),
\end{equation}
where \(\mathcal{E}_g\) denotes the guidance encoder, \(S\) is the number of decoding stages, and \(f_g^{s}\) denotes the guidance feature at stage \(s\).

Hybrid guidance maps can directly provide additional texture and semantic information. More importantly, they introduce physically meaningful constraints, such as geometry cues and prior material predictions, as well as semantic constraints, such as the correspondence between body structure and material category. These complementary constraints guide the decoder toward a more reasonable solution space, thereby improving estimation accuracy.

However, directly fusing all guidance features with the input features is suboptimal. Such a strategy tends to cause two failure modes: either semantic information is weakened because texture details are overemphasized, or texture information is suppressed because semantic guidance dominates the fusion. As a result, the model cannot fully exploit the complementary guidance information. The key to effective guidance therefore lies in designing a fusion strategy that allows texture-level and semantic-level cues to guide material decoding at the stages where they are most informative.

To this end, we propose the \FusionFullLow~(\FusionFullShort) to better exploit the heterogeneous guidance maps. Specifically, we first concatenate the two features along the channel dimension:
\begin{equation}
	f_{cat}^s = [f_g^s ; f_d^s],
\end{equation}
and feed it into a lightweight convolutional weighting network \(\phi^s(\cdot)\) to predict spatially varying fusion weights:
\begin{equation}
	[w_g^s, w_d^s] = \mathrm{Softmax}(\phi^s(f_{cat}^s)),
\end{equation}
where \(\phi^s(\cdot)\) consists of stacked convolution, normalization, and activation layers followed by a $1\times1$ convolution that outputs two channels corresponding to guidance and decoder weights. 
The softmax operation is applied across the two channels at each spatial location to ensure normalized and complementary weighting.

The fused feature is then computed as:
\begin{equation}
	f_{out}^s = w_g^s \odot f_g^s + w_d^s \odot f_d^s,
\end{equation}
where \(\odot\) denotes element-wise multiplication.
This design enables spatially adaptive fusion, allowing the network to dynamically emphasize guidance or decoder features according to local content.

Following DIC~\cite{DIC_CVPR_2025}, we adopt sparse fusion to avoid feature redundancy. Specifically, we fuse the guidance features at three stages: shallow, medium, and deep. The adaptive feature fusion (its architecture depicted in Fig.~\ref{img-guidance_model}) is used for shallow- and medium-level feature fusion, while a Transformer~\cite{Transformer} is employed for deep-layer feature fusion to better model high-level semantic dependencies. 
The guidance encoder consists of eight down-sampling blocks with channel dimensions of (64, 128, 128, 128, 512, 512, 1024) and (1024), respectively. Each block includes a 3x3 convolution, group normalization, and ReLU activation function. The corresponding feature map sizes are (1024, 512, 256, 128, 64, 32, 16), and (8). Among these, the block with a feature map size of 512 is referred to as the shallow layer, the one with a feature map size of 64 is the medium layer, and the block with a feature map size of 8 is considered the deep layer.
The structures of the three decoders are identical and symmetric to the encoder structure.
 
The Transformer-based fusion is implemented by a multi-head attention module $\mathrm{MHA}(\cdot)$, with 8 attention heads.
Specifically, given the current decoder feature \(x \in \mathbb{R}^{B \times C \times H \times W}\) and the corresponding guidance feature \(f_g^{s_{\text{deep}}} \in \mathbb{R}^{B \times C \times H \times W}\), we first flatten the spatial dimensions and transpose the features into token sequences:
\begin{equation}
	\begin{aligned}
		q &= \mathrm{Flatten}(x) \in \mathbb{R}^{B \times HW \times C}, \\
		k &= \mathrm{Flatten}(f_g^{s_{\text{deep}}}) \in \mathbb{R}^{B \times HW \times C}, \\
		v &= k, \\
		f_{\text{attn}} &= \mathrm{MHA}(q, k, v), \\
		\tilde{x} &= \mathrm{Reshape}(f_{\text{attn}}) \in \mathbb{R}^{B \times C \times H \times W}.
	\end{aligned}
\end{equation}

This deep-layer feature fusion allows texture-dominant and semantic-dominant cues to guide material decoding at appropriate levels, thereby improving both estimation accuracy and physical plausibility.
To better match the properties of different materials, we use three decoders (normal decoder \(\mathcal{D}_{n}\), diffuse decoder \(\mathcal{D}_{d}\), and SSR decoder \(\mathcal{D}_{ssr}\)) to estimate material maps.

In Fig.~\ref{img-guidance_model}, we visualize the feature fusion weight maps of the three material decoders at the shallow (\(w_{g}^{s}\)) and middle (\(w_{g}^{m}\)) layers. We observe that: (1) for guidance maps mainly containing texture features, such as ``P-Nor", the weights of the ``Nor" decoder at the shallow layer are significantly higher than those at the middle layer; (2) for guidance maps mainly containing semantic features, such as ``Input" and ``Densepose", the ``Diff" and ``SSR" decoders follow a consistent pattern, where the weights at the middle layer are higher than those at the shallow layer. These observations suggest that \FusionFullShort ~enables the model to distinguish between different types of guidance features and dynamically allocate hierarchical fusion weights according to feature characteristics, thus making more effective use of heterogeneous guidance information.

\subsection{Loss Function}
During training, the overall loss \(\mathcal{L}_{total}\) of our model consists of a material loss \(\mathcal{L}_{material}\) and a rendering loss \(\mathcal{L}_{relit}\):
\begin{equation}
	\label{loss_total}
	\mathcal{L}_{total} = \mathcal{L}_{material} + \mathcal{L}_{relit},
\end{equation}
where \(\mathcal{L}_{material}\) enforces pixel-wise consistency between predicted and ground-truth material maps, and \(\mathcal{L}_{relit}\) encourages rendering consistency under multiple lighting conditions.

The material loss consists of normal loss \(\mathcal{L}_{n}\), diffuse albedo loss \(\mathcal{L}_{d}\), roughness loss \(\mathcal{L}_{r}\), specular albedo loss \(\mathcal{L}_{s}\), and subsurface scattering loss \(\mathcal{L}_{sss}\). Each term is computed using a pixel-wise L1 loss between the prediction and the ground truth. Therefore, the material loss is defined as:
\begin{equation}
	\label{loss_material}
	\mathcal{L}_{material} = \mathcal{L}_{n} + \mathcal{L}_{d} + \mathcal{L}_{r} + \mathcal{L}_{s} + \mathcal{L}_{sss}.
\end{equation}

Rendering the appearance under various illuminations and calculating the corresponding loss help disentangle material from lighting, thereby improving the robustness of material estimation. Following \HuMat~\cite{HumanMaterial}, we compute the rendering loss under 37 lighting conditions with different light positions. The multi-illumination rendering loss \(\mathcal{L}_{relit}\) is defined as:
%\begin{equation}
%	\label{loss_relighting}
%	\begin{split}
%		\mathcal L_{relit}=&\frac{1}{M}\sum_{i=1}^{M} func(\mathcal{R}(gt, light_{i}), \mathcal{R}(pred, light_{i})),
%	\end{split}
%\end{equation}
%where \(func\) denotes the combination of L1 loss and VGG loss~\cite{VGG_loss}, \(pred\) and \(gt\) denote the predicted and ground-truth materials, respectively, \(\mathcal{R}\) is the PBR shader implemented with the Disney BSDF model~\cite{DisneyBRDF_2012_burley}, \(light_{i}\) denotes the illumination used for rendering, and \(M\) (set to 37) represents the number of point lights distributed on the positive hemisphere.
\begin{equation}
	\label{loss_relighting}
	\mathcal L_{relit}
	=
	\frac{1}{M}\sum_{i=1}^{M}
	\ell\big(\mathcal{R}(gt, light_{i}), \mathcal{R}(pred, light_{i})\big),
\end{equation}
where \(\ell\) denotes the combination of L1 loss and VGG loss~\cite{VGG_loss}, \(pred\) and \(gt\) denote the predicted and ground-truth materials, respectively, \(\mathcal{R}\) is the PBR shader implemented with the Disney BSDF model~\cite{DisneyBRDF_2012_burley}, \(light_i\) denotes the illumination used for rendering, and \(M\) (set to 37) is the number of point lights distributed over the upper hemisphere.

\subsection{Rendering Function}
To provide rendering supervision during training, we use a physics-based shader to synthesize appearance from material maps and lighting:
\begin{equation}
	\label{reflection_equation}
	{L}(x) =  \int_{\Omega} L_{i} \cdot f_{r}(materials,{w}_{i},{w}_{o}) \cdot ({w}_{i} \cdot {n}) \cdot {dw}_{i},
\end{equation}
where \(L(x)\) is the rendered radiance at surface point \(x\), \(L_i\) is the incident illumination, and \(f_r\) denotes the BSDF. We adopt the Disney BSDF model~\cite{DisneyBRDF_2012_burley}, which supports subsurface scattering and is well suited for modeling human skin. The material parameters include normal (\(n\)), diffuse albedo (\(d\)), roughness (\(r\)), specular albedo (\(s\)), and subsurface scattering (\(sss\)), while \(w_i\) and \(w_o\) are the incident and outgoing light directions.

\textbf{Disney BSDF} is a physical model proposed by Disney Research. It aims to efficiently and realistically simulate the optical properties of various materials and is widely used in the field of material rendering in computer graphics.
Disney BSDF supports a large number of advanced parameters, such as subsurface scattering, metallic, clearcoat and anisotropy. To meet the needs of our work, we use subsurface scattering to enhance the rendering realism of human skin in addition to the common normal, diffuse albedo, roughness and specular albedo.
Therefore, BSDF ($f_r$) is calculated by:
\begin{equation}
	f_r = \underbrace{
		(1-sss) \cdot F_{baseDiff}+ sss \cdot F_{sss}
	}_{\text{diffuse}} + \underbrace{\frac{F_m D_m G_m}{4 |n \cdot \omega_{i}|}}_{\text{specular}}.
\end{equation}
$F_{baseDiff}$ can be calculated by:
\begin{equation}
	F_{baseDiff} = \frac{\text{d}}{\pi} F_D(\omega_{i}) F_D(\omega_{o}) |n \cdot \omega_{o}|,
\end{equation}
where $d$ is diffuse albedo,
$F_D(\omega)$ can be calculated by:
\begin{equation}
	F_D(\omega) = \left(1 + (F_{D90} - 1)(1 - |n \cdot \omega|)^5\right),
\end{equation}
where $\omega$  is the light or camera vector. 
$F_{D90}$ can be calculated by:
\begin{equation}
	F_{D90} = \frac{1}{2} + 2 \cdot r \cdot |h \cdot \omega_{o}|^2,
\end{equation}
where $r$ is roughness, $h$ is the half vector. 
$h$ is calculated by:
\begin{equation}
	h = \frac{\omega_{i} + \omega_{o}}{|\omega_{i} + \omega_{o}|}.
\end{equation}

We use a BRDF approximation of the subsurface scattering by modifying the Lommel-Seeliger law:
\begin{equation}
	\begin{split}
		f_{\text{sss}} &= \frac{1.25 \cdot \text{d}}{\pi} \left( F_{SS}(\omega_{i}) F_{SS}(\omega_{o}) \right. \\
		&\quad \left. \left( \frac{1}{|n \cdot \omega_{i}| + |n \cdot \omega_{o}|} - 0.5 \right) + 0.5 \right) |n \cdot \omega_{o}|,
	\end{split}
\end{equation}
where 
\begin{equation}
	\begin{split}
		F_{SS}(\omega) &= \left( 1 + (F_{SS90} - 1)(1 - |n \cdot \omega|)^5 \right), \\
		F_{SS90} &= \text{roughness} \cdot |h \cdot \omega_{o}|^2.
	\end{split}
\end{equation}

For specular reflection, Burley uses a standard Cook-Torrance microfacet BRDF. 
$F_m$ is the Fresnel reflection, we adapt the Schlick approximation:
\begin{equation}
	F_m = \text{d} + (1 - \text{d})(1 - |h \cdot \omega_{o}|)^5.
\end{equation}

$D_m$ is the probability density of the distribution of a microfacet normal, and it can be calculated by:
\begin{equation}
	D_m = \frac{1}{\pi \max(0.0001, r^2) \left( \frac{h^2}{\max(0.0001, r^2)^2} \right)^2}.
\end{equation}

$G_m$ is the masking-shadowing term. Before computing $G_m$, we apply the roughness remapping:
\begin{equation}
	r = \frac{\text{r} + 1}{2},
\end{equation}
where $r$ is the remapped roughness used only in the calculation of shadowing and masking. Then,
\begin{equation}
	G_m = G(\omega_{i}) G(\omega_{o}),
\end{equation}
where 
\begin{equation}
	\begin{split}
		G(\omega) &= \frac{1}{1 + \Lambda(\omega)}, \\
		\Lambda(\omega) &= \frac{\sqrt{1 + (\omega_l \cdot \max(0.0001, r^2))^2 - 1}}{2}.
	\end{split}
\end{equation}

\begin{figure*}[!t]
  \centering % 居中
  \begin{minipage}[b]{\linewidth} % 整体划分一个页面
  \centering % 居中
  \newcommand{\myvspace}{1.0 pt} % 定义行之间的空隙
  \newcommand{\widthOfFullPage}{1} % 1 / 列数
  \newcommand{\widthOfMiniPage}{0.98}
  \newcommand{\format}{png}
  \includegraphics[width=\widthOfMiniPage\linewidth]{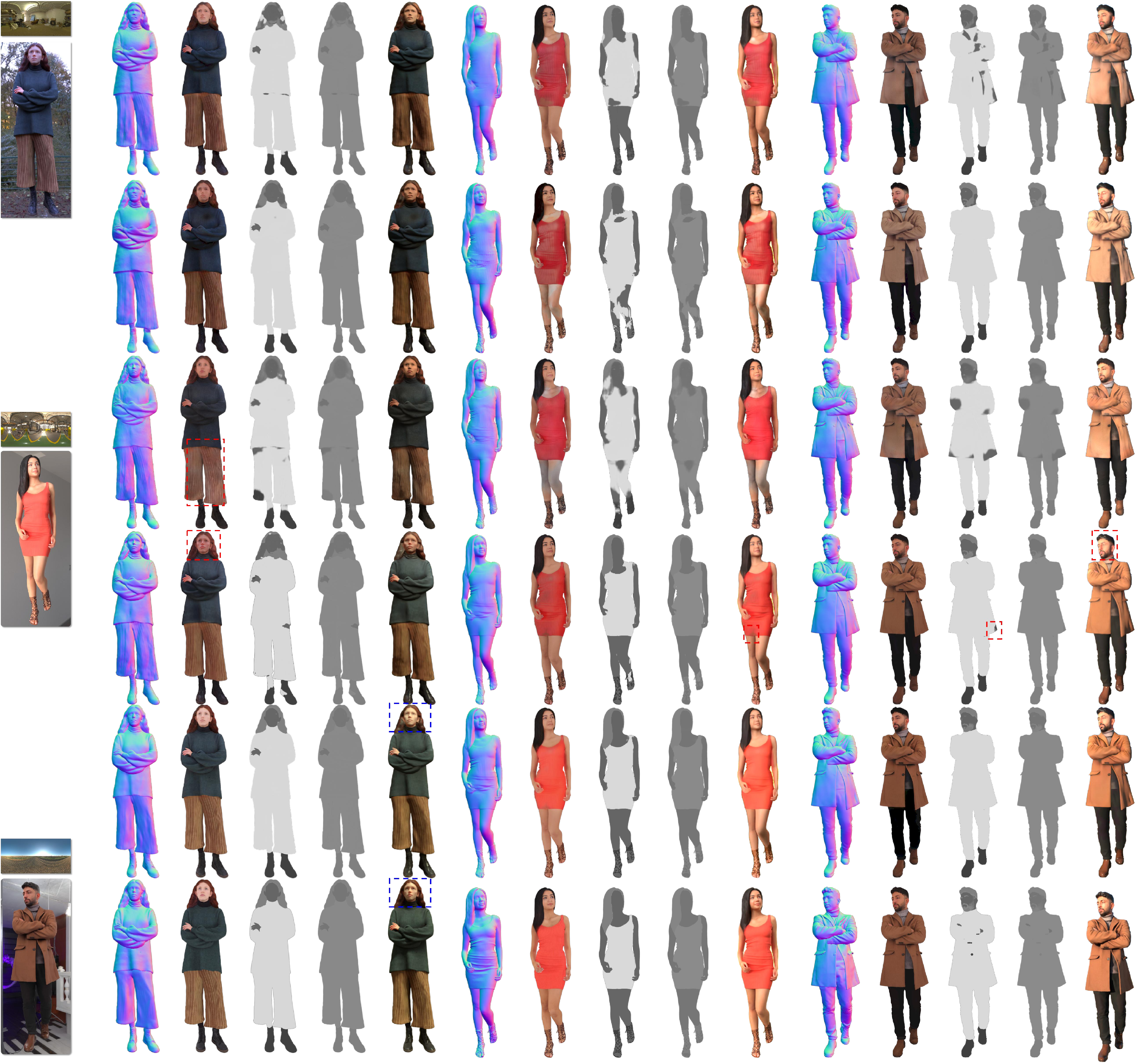}
  \begin{picture}(0,0)
  	\small
  	\put(-450,419){\rotatebox{90}{\makebox(0,0)[c]{\RADN}}}
  	\put(-450,343){\rotatebox{90}{\makebox(0,0)[c]{\HATSNet}}}
  	\put(-450,266){\rotatebox{90}{\makebox(0,0)[c]{\AFHR}}}
  	\put(-450,192){\rotatebox{90}{\makebox(0,0)[c]{\HuMat}}}
  	\put(-450,118){\rotatebox{90}{\makebox(0,0)[c]{Ours}}}
  	\put(-450,42){\rotatebox{90}{\makebox(0,0)[c]{GT}}}
  \end{picture}  
  \begin{flushleft}
    \small
    \vspace{-0.5em}
    \hspace{1.0em}Input
    \hspace{3.0em}N
    \hspace{2.3em}D
    \hspace{2.2em}R
    \hspace{2.3em}S
    \hspace{1.7em}Relit
    \hspace{1.4em}N
    \hspace{2.1em}D
    \hspace{2.2em}R
    \hspace{2.4em}S
    \hspace{1.7em}Relit
    \hspace{1.6em}N
    \hspace{2.3em}D
    \hspace{2.3em}R
    \hspace{2.3em}S
    \hspace{1.3em}Relit
  \end{flushleft}
  \end{minipage}
\caption{
	Materials estimation and relighting comparison with previous works (\RADN~\cite{RADN_2018_TOG_single_image_natural_svbrdf}, \HATSNet~\cite{HATSNet_2021_TOG_single_image_natural_svbrdf}, \AFHR\cite{AFHR_2025_EGSR}, and \HuMat~\cite{HumanMaterial}) on \HumatDataset dataset. ``Input" means the input image and a HDR map for relighting.
	``N" means normal, ``D" means diffuse albedo, ``R" means roughness, ``S" means specular albedo, ``SSS" means subsurface scattering, and ``Relit" means relighting image.
}
\label{img-material_comparision_on_syn}
\end{figure*}
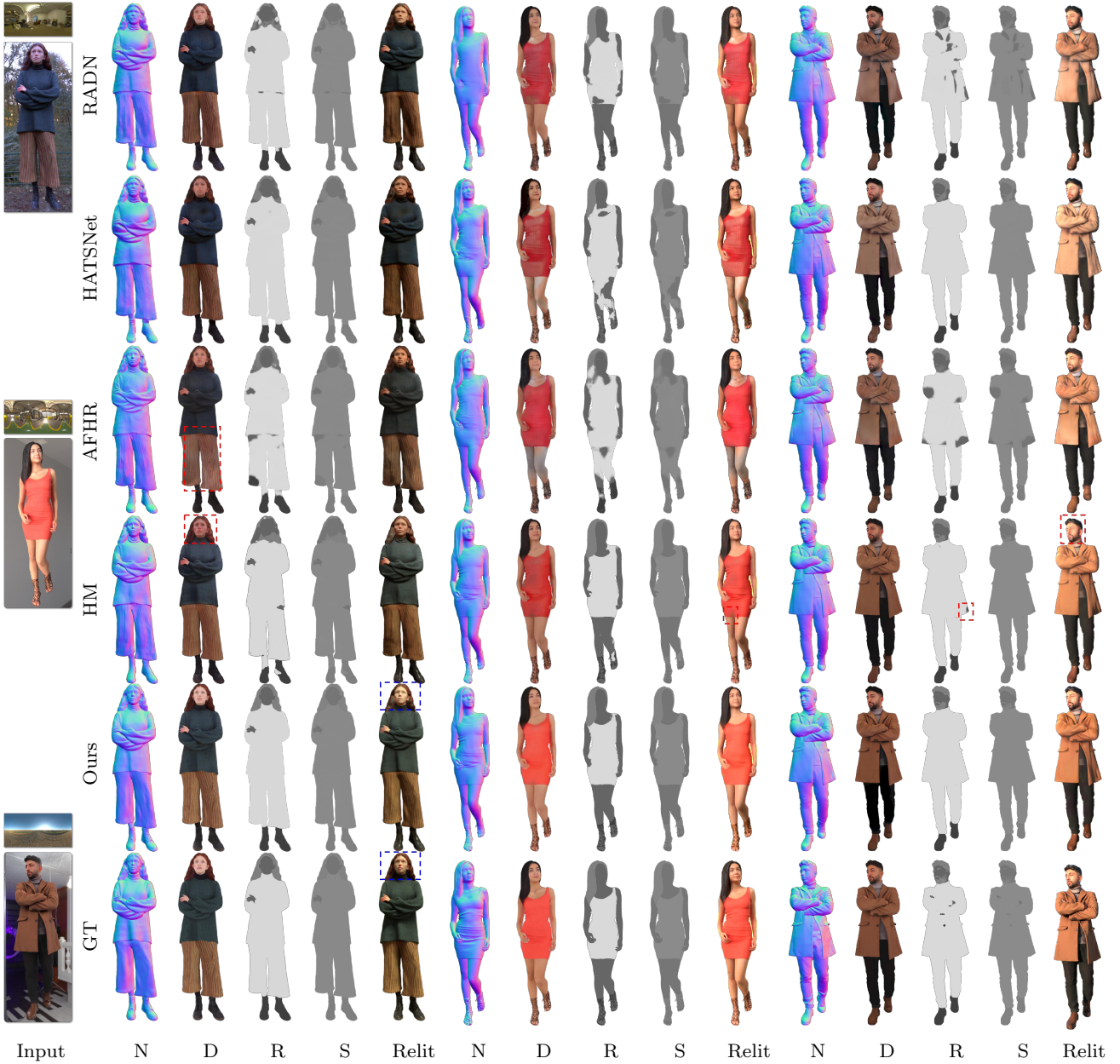
\begin{table*}[!ht]
	\centering
	\caption{
		Materials and rendering result error (in terms of PSNR) comparison between previous works 
		on \HumatDataset dataset.  
		``Point" means relighting under the fixed single point light. ``Real" and ``Syn" means relighting under four real and one synthetic HDR environment maps, respectively.
	}
	\scalebox{0.9}{
		\begin{tblr}{
				row{1} = {c},
				cell{1}{1} = {r=3}{c},
				cell{1}{2} = {c=5}{c},  % 原6列减1列
				cell{1}{7} = {c=6}{c},  % 原8列减1列
				cell{1}{13} = {c=3}{c}, % 原14列减1列
				cell{2}{2} = {r=2}{c},
				cell{2}{3} = {r=2}{c},
				cell{2}{4} = {r=2}{c},
				cell{2}{5} = {r=2}{c},
				cell{2}{6} = {r=2}{c},  % 原7列减1列
				cell{2}{7} = {r=2}{c},  % 原8列减1列
				cell{2}{8} = {c=4}{c},  % 原9列减1列
				cell{2}{12} = {r=2}{c}, % 原13列减1列
				cell{2}{13} = {r=2}{c}, % 原14列减1列
				cell{2}{14} = {r=2}{c}, % 原15列减1列
				cell{2}{15} = {r=2}{c}, % 原16列减1列
				% 单元格居中设置（简化重复项）
				cell{4-10}{2-9,13,15} = {c},  % 行数范围从4-9扩展为4-10
				hline{1,11} = {-}{0.13em},  % 底线从10调整为11
				% 调整水平线分割逻辑
				hline{2} = {2}{l}, hline{2} = {3-5}{},hline{2} = {6}{r},
				hline{2} = {7}{l},hline{2} = {8-11}{},hline{2} = {12}{r},
				hline{2} = {13}{l},hline{2} = {14-15}{},hline{2} = {16}{r},
				hline{3} = {8}{l}, hline{3} = {9-10}{}, hline{3} = {11}{r},
				hline{4} = {-}{0.1em},
			}
			Method                                        & Material &   &   &   &     & Relighting &         &         &         &         &     & Mean      &            &       \\
			& N         & D & R & S & SSS & Point     & Real    &         &         &         & Syn & Material & Relit.& Total \\
			&           &   &   &   &     &           & Real\_1 & Real\_2 & Real\_3 & Real\_4 &     &           &            &       \\
			RADN\cite{RADN_2018_TOG_single_image_natural_svbrdf}                                           &20.3          &26.0   &22.4   &38.1   &/     &21.4           &21.5         &21.9         &21.7         &21.8         &22.6     &26.7           &21.8            &24.3       \\  
			\HATSNet \cite{HATSNet_2021_TOG_single_image_natural_svbrdf}        &20.5    &26.2  &22.9   &38.4   & /            &21.9   &21.7     &21.9         &21.8        &22.0     &22.8     &27.0   &22.0 &24.5 \\
			\AFHR\cite{AFHR_2025_EGSR}   &20.8   &26.0   &22.5      & 38.2        & /          & /        &21.6     &22.0         &21.7        &22.5     &23.6     &26.9   &22.3  &24.6 \\
			\FBHR\cite{FBHR_2021_EGSR}               & /         &25.7   & /        & /        & /           & /        &21.7     &22.7         &22.4        &22.1     &20.5     &25.7   &21.9  &23.8 \\
			\TotalReLighting\cite{TotalRelighting_2021_Sig_single_image_human_phong_neural_render} &19.5    &25.9   & /        & /        & /           & /        &22.0    &\underline{23.1}         &22.6        &22.3     &20.7     &22.7   &22.1  &22.4 \\		
			\HuMat\cite{HumanMaterial}                &\underline{21.7}   &\underline{27.2}   &\underline{24.7}  &\underline{40.6}   &\underline{42.0} &\underline{24.0}   &\underline{22.4}     &22.8        &\underline{22.8}        &\underline{22.4}    &\underline{24.5}     &\underline{31.2}   &\underline{23.2}   &\underline{27.2}    \\
			Ours              &\textbf{22.5}          &\textbf{28.3}   &\textbf{25.3}   &\textbf{40.8}   &\textbf{42.5}     &\textbf{24.8}           &\textbf{22.9}         &\textbf{23.2}         &\textbf{23.1}         &\textbf{22.8}         &\textbf{24.9}     &\textbf{31.9}           &\textbf{23.6}            &\textbf{27.7}       
		\end{tblr}
	}
	\label{Table-comparison_on_openhumanbrdf_dataset_test}
\end{table*}

\begin{figure*}[!ht]
	\centering % 居中
	\begin{minipage}[b]{\linewidth} % 整体划分一个页面
		\centering % 居中
		\newcommand{\myvspace}{1.0 pt} % 定义行之间的空隙
		\newcommand{\widthOfFullPage}{1} % 1 / 列数
		\newcommand{\widthOfMiniPage}{0.99}
		\newcommand{\format}{png}
		\subfloat{
		\begin{minipage}[b]{\widthOfFullPage\linewidth} % 继续划分子页面
			\centering
			\includegraphics[width=\widthOfMiniPage\linewidth]{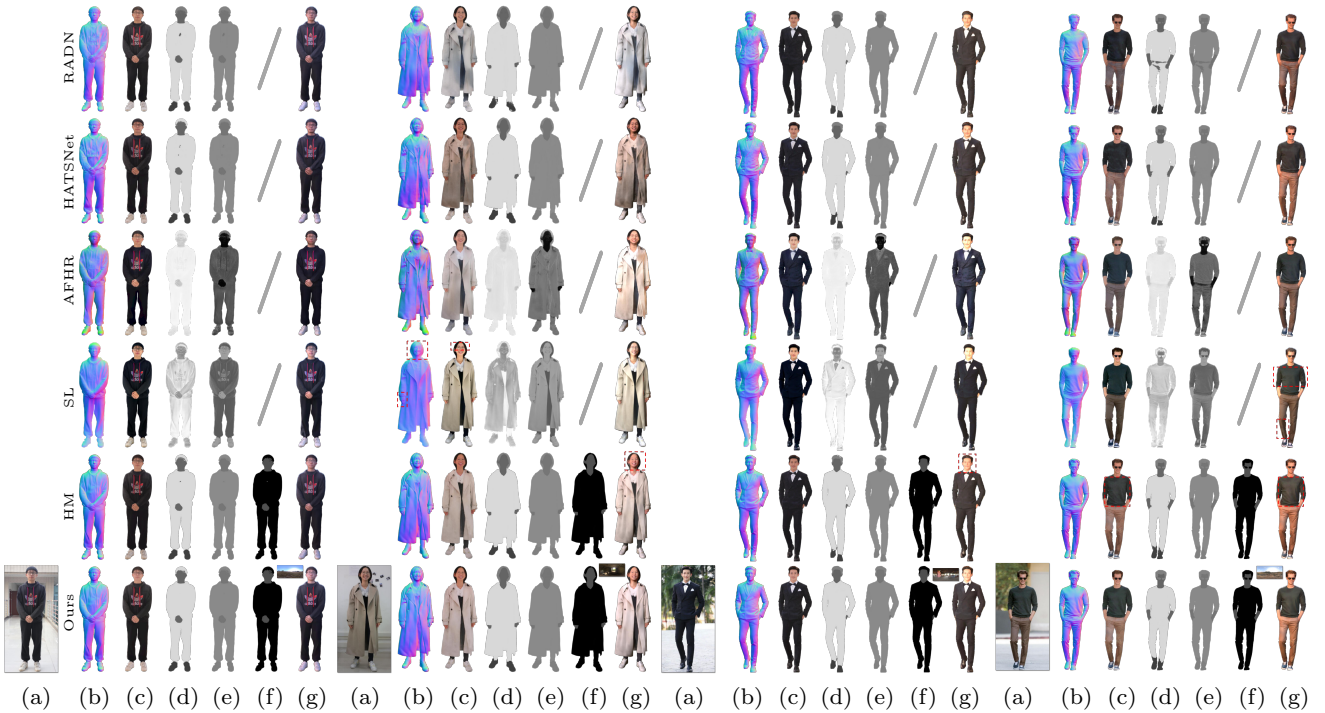}
			\begin{picture}(0,0)
				\tiny
				\put(-469,233){\rotatebox{90}{\makebox(0,0)[c]{\RADN}}}
				\put(-469,190){\rotatebox{90}{\makebox(0,0)[c]{\HATSNet}}}
				\put(-469,149){\rotatebox{90}{\makebox(0,0)[c]{\AFHR}}}
				\put(-469,105){\rotatebox{90}{\makebox(0,0)[c]{\SwitchLight}}}
				\put(-469,64){\rotatebox{90}{\makebox(0,0)[c]{\HuMat}}}
				\put(-469,23){\rotatebox{90}{\makebox(0,0)[c]{Ours}}}
			\end{picture}
			\end{minipage}
			}
		\begin{flushleft}
		\small
		\vspace{-0.5em}
		\hspace{1.1em}(a)
		\hspace{0.8em}(b)
		\hspace{0.3em}(c)
		\hspace{0.2em}(d)
		\hspace{0.2em}(e)
		\hspace{0.3em}(f)
		\hspace{0.2em}(g)
		\hspace{0.6em}(a)
		\hspace{0.6em}(b)
		\hspace{0.3em}(c)
		\hspace{0.2em}(d)
		\hspace{0.2em}(e)
		\hspace{0.3em}(f)
		\hspace{0.2em}(g)
		\hspace{0.6em}(a)
		\hspace{0.8em}(b)
		\hspace{0.3em}(c)
		\hspace{0.2em}(d)
		\hspace{0.2em}(e)
		\hspace{0.3em}(f)
		\hspace{0.2em}(g)
		\hspace{0.6em}(a)
		\hspace{0.8em}(b)
		\hspace{0.3em}(c)
		\hspace{0.2em}(d)
		\hspace{0.2em}(e)
		\hspace{0.3em}(f)
		\hspace{0.2em}(g)
	\end{flushleft}
\end{minipage}
\caption{
Materials estimation and relighting comparison with previous works on the real data. 
(a) Input. (b) Normal. (c) Diffuse albedo. (d) Roughness. (e) Specular albedo. (f) Subsurface scattering. (g) Relighting.
}
\label{img-material_comparision_on_real_3}
\end{figure*}
\begin{table*}[!ht]
	\centering
	\caption{
		Materials and rendering result error (in terms of PSNR) of ablation study on \HumatDataset dataset.
	}
	\scalebox{0.9}{
		\begin{tblr}{
				row{1} = {c},
				cell{1}{1} = {r=3}{c},
				cell{1}{2} = {c=5}{c},
				cell{1}{7} = {c=6}{c},
				cell{1}{13} = {c=3}{c},
				cell{2}{2} = {r=2}{c}, 
				cell{2}{3} = {r=2}{c},
				cell{2}{4} = {r=2}{c},
				cell{2}{5} = {r=2}{c},
				cell{2}{6} = {r=2}{c},
				cell{2}{7} = {r=2}{c},
				cell{2}{8} = {c=4}{c},
				cell{2}{12} = {r=2}{c},
				cell{2}{13} = {r=2}{c},
				cell{2}{14} = {r=2}{c},
				cell{2}{15} = {r=2}{c},
				cell{3-11}{2-15} = {c},
				hline{1,11} = {-}{0.13em},
				hline{2} = {2}{l}, hline{2} = {3-5}{},hline{2} = {6}{r},
				hline{2} = {7}{l},hline{2} = {8-11}{},hline{2} = {12}{r},
				hline{2} = {13}{l},hline{2} = {14-15}{},hline{2} = {16}{r},
				hline{3} = {8}{l}, hline{3} = {9-10}{}, hline{3} = {11}{r},
				hline{4} = {-}{0.1em},
			}
			Method    & Material &   &   &   &     & Relighting &         &         &         &         & Syn & Mean      &            &       \\
			& N         & D & R & S & SSS & Point     & Real    &         &         &         &     & Material & Relit.& Total \\
			&           &   &   &   &     &           & Real\_1 & Real\_2 & Real\_3 & Real\_4 &     &           &            &       \\
			w/o Guid   &21.7 &26.2   &23.4   &38.1   &40.3  &23.3  &22.3  &22.6  &22.7  &22.2  &24.2  &29.9  &22.9 &26.4  \\
			w/o P-Guid &22.0 &\underline{27.6}   &\underline{25.0}   &\underline{40.6}   &\underline{42.3}  &\underline{24.4}  &\underline{22.7}  &\underline{22.9}  &\underline{23.0}  &\underline{22.7}  &\underline{24.8} &\underline{31.5}  &\underline{23.4} &\underline{27.5}  \\
			w/o \FusionFullShort  &21.7 &26.1   &24.3   &39.7   &41.8  &23.8  &22.4  &22.7  &22.7  &22.3  &24.0 &30.7  &23.0 &26.9 \\
			w/o \Dino   & 22.1         &27.4   &24.8   &40.2   &42.3     &24.0           &22.6         &22.7         &22.8         &22.6         &24.5     &31.4           &23.2            &27.3       \\  
			w/o \HuMat   &22.0          &26.6   &24.1   &39.1   &41.7     &23.6           &22.3         &22.6         &22.6         &22.2         &24.3     &30.7           &22.9            &26.8       \\  
			w/o \AFHR   &\underline{22.1}          &27.1   &24.9   &40.3   &\underline{42.3}     &24.0           &22.6         &22.1         &22.9         &22.5         &24.4     &31.3           &23.1            &27.2       \\  
			Ours              &\textbf{22.5}          &\textbf{28.3}   &\textbf{25.3}   &\textbf{40.8}   &\textbf{42.5}     &\textbf{24.8}           &\textbf{22.9}         &\textbf{23.2}         &\textbf{23.1}         &\textbf{22.8}         &\textbf{24.9}     &\textbf{31.9}           &\textbf{23.6}            &\textbf{27.7}         
		\end{tblr}
	}
	\label{table-ablation_on_syn}
\end{table*}

\subsection{Implementation Details}

\textbf{Evaluation Metrics.}
To quantitatively evaluate the material estimation performance of our method and the comparative methods, we calculate the PSNR between the estimated material maps and the ground-truth material maps, as well as the PSNR of the rendered results under different illumination settings (e.g., point lights, real environment maps, and synthetic environment maps).

\textbf{Training Details.}
Our method is trained on four NVIDIA RTX 3090 GPUs (24GB memory each) with a batch size of 1 for a maximum of 100 epochs, which takes approximately two days. We employ the AdamW~\cite{AdamW_ICLR_2017} optimizer with default settings and a learning rate of \(1\times10^{-4}\).
The DinoV2 version is ``ViT-B/14 distilled".

\begin{figure*}[!ht]
  \centering % 居中
  \begin{minipage}[b]{\linewidth} % 整体划分一个页面
  \centering % 居中
  \newcommand{\myvspace}{1.0 pt} % 定义行之间的空隙
  \newcommand{\widthOfFullPage}{1} % 1 / 列数
  \newcommand{\widthOfMiniPage}{1.0}
  \newcommand{\format}{png}
  \includegraphics[width=\widthOfMiniPage\linewidth]{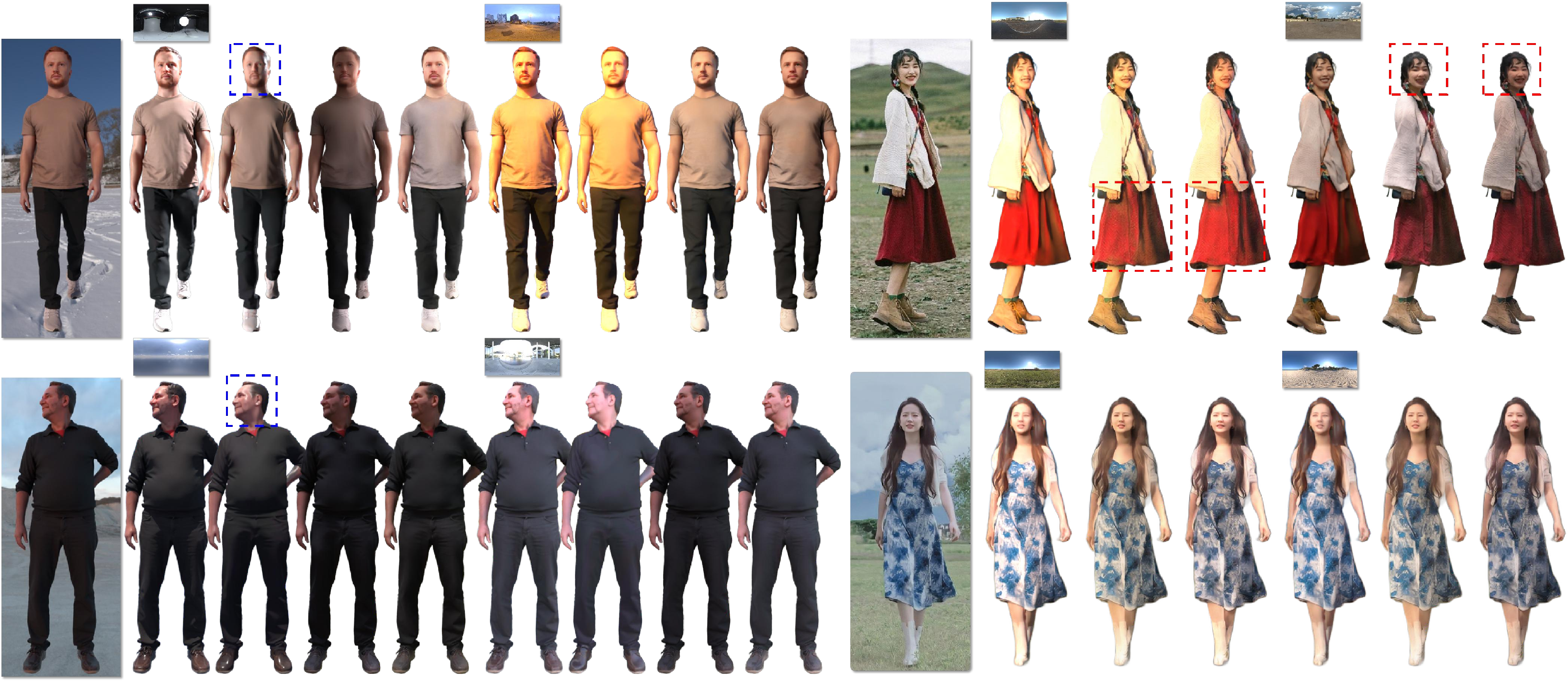}
  \begin{flushleft}
  	\small
  	\vspace{-0.5em}
  	\hspace{1.2em}Input
    \hspace{1.4em}GT
    \hspace{0.8em}Ours
    \hspace{0.2em}\FBHR
    \hspace{0.8em}\TotalReLighting
    \hspace{1.4em}GT
    \hspace{0.8em}Ours
    \hspace{0.2em}\FBHR
    \hspace{0.8em}\TotalReLighting
    \hspace{2.1em}Input
    \hspace{1.5em}Ours
    \hspace{0.7em}\FBHR
    \hspace{1.1em}\TotalReLighting
    \hspace{1.2em}Ours
    \hspace{0.7em}\FBHR
    \hspace{1.0em}\TotalReLighting
  \end{flushleft}
  \end{minipage}
\caption{
Relighting comparison with previous works (\FBHR~\cite{FBHR_2021_EGSR} and \TotalReLighting~\cite{TotalRelighting_2021_Sig_single_image_human_phong_neural_render}).
The left two are the results on \HumatDataset dataset, and the right two are the results on real data.
}
\label{img-relighting_comparision}
\end{figure*}
\begin{figure}[!ht]
  \centering % 居中
  \begin{minipage}[b]{\linewidth} % 整体划分一个页面
  \centering % 居中
  \newcommand{\myvspace}{1.0 pt} % 定义行之间的空隙
  \newcommand{\widthOfFullPage}{1} % 1 / 列数
  \newcommand{\widthOfMiniPage}{1.0}
  \subfloat{
    \begin{minipage}[b]{\widthOfFullPage\linewidth} % 继续划分子页面
      \centering
    \includegraphics[width=\widthOfMiniPage\linewidth]{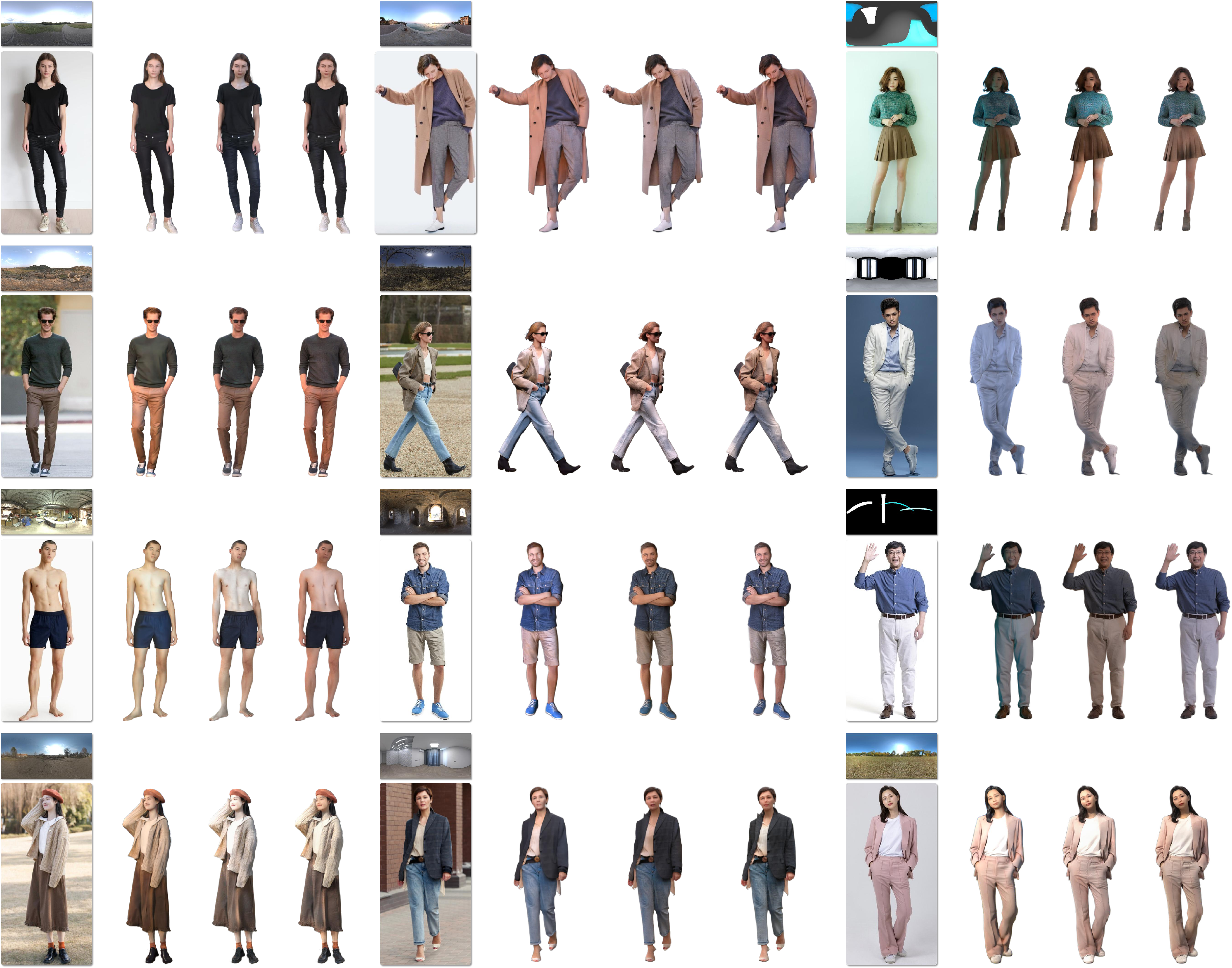}\vspace{\myvspace}   
    \end{minipage}
  }
  \begin{flushleft}
  	\small
%  	\tiny
  	\vspace{-0.5em}
  	\hspace{0.7em}(a)
    \hspace{0.7em}(b)
    \hspace{0.3em}(c)
    \hspace{0.2em}(d)
    \hspace{0.7em}(a)
    \hspace{0.7em}(b)
    \hspace{0.7em}(c)
    \hspace{0.7em}(d)
    \hspace{1.4em}(a)
    \hspace{0.7em}(b)
    \hspace{0.3em}(c)
    \hspace{0.2em}(d)
  \end{flushleft}
  \end{minipage}
\caption{
More relighting comparison on the real data. (a) Input image and HDR map. (b) Ours. (c) \FBHR~\cite{FBHR_2021_EGSR}. (d) \TotalReLighting~\cite{TotalRelighting_2021_Sig_single_image_human_phong_neural_render}.
}
\label{img-more_relighting_comparision}
\end{figure}
\section{Experiments}
\label{sec:experiments}
\subsection{Experimental Setup}
\textbf{Datasets.}
The \HumatDataset dataset provides human PBR material data, so we conduct supervised training on it.
This dataset contains 147 sets of human data. 127 of them are used for training, and 20 are used for testing. For the training and testing data, this dataset renders 100 and 10 multi-view image sets for each model, respectively.
The resolution of each image set is $512\times512$. Each set includes the foreground mask, PBR material maps (normal, diffuse albedo, roughness, specular albedo, and subsurface scattering), and five relighting results under five (four real and one synthetic) HDR environment maps.

\subsection{Comparisons with State-of-the-art Methods}
We compare \Ours with representative state-of-the-art methods from two related tasks: PBR material estimation and human relighting. The PBR material estimation baselines include \RADN~\cite{RADN_2018_TOG_single_image_natural_svbrdf}, \HATSNet~\cite{HATSNet_2021_TOG_single_image_natural_svbrdf}, \SwitchLight~\cite{SwitchLight_2024_CVPR_pbr_and_neural_render}, \AFHR~\cite{AFHR_2025_EGSR}, and \HuMat~\cite{HumanMaterial}. The relighting baselines include \FBHR~\cite{FBHR_2021_EGSR} and \TotalReLighting~\cite{TotalRelighting_2021_Sig_single_image_human_phong_neural_render}. We conduct comparisons on both synthetic and real data whenever the corresponding methods are available.
Since \SwitchLight~\cite{SwitchLight_2024_CVPR_pbr_and_neural_render} only provides an online testing interface, we compare with it only on real images to ensure a fair evaluation.

\textbf{Quantitative Comparison.}
As shown in Table~\ref{Table-comparison_on_openhumanbrdf_dataset_test}, \Ours achieves the best overall performance on both material estimation and relighting. Compared with existing methods, our approach consistently improves the reconstruction quality of the estimated material maps, while also leading to more faithful relighting results under novel illumination. These results demonstrate that the proposed hybrid guidance maps effectively reduce the ambiguity of single-image human material estimation, and that \FusionFullShort ~further improves the use of heterogeneous guidance cues during decoding.

Among all compared methods, \HuMat is the strongest baseline in our setting, yet \Ours still achieves further gains on most metrics, especially for diffuse albedo. This suggests that, beyond using the input image as a guidance signal, introducing hybrid priors and stage-wise adaptive fusion is crucial for improving appearance decomposition. In addition, our method remains competitive on geometry- and reflectance-related maps, including normal, roughness, specular albedo, and subsurface scattering, indicating that the proposed framework generalizes well across different material attributes.

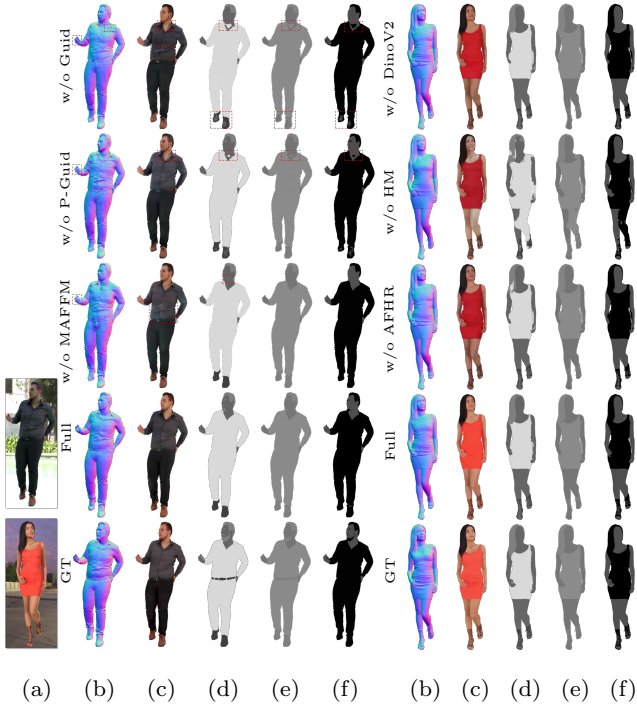
\begin{figure}[!ht]
  \centering % 居中
  \begin{minipage}[b]{\linewidth} % 整体划分一个页面
  \centering % 居中
  \newcommand{\myvspace}{1.0 pt} % 定义行之间的空隙
  \newcommand{\widthOfFullPage}{1} % 1 / 列数
  \newcommand{\widthOfMiniPage}{0.99}
  \newcommand{\format}{png}
  \includegraphics[width=\widthOfMiniPage\linewidth]{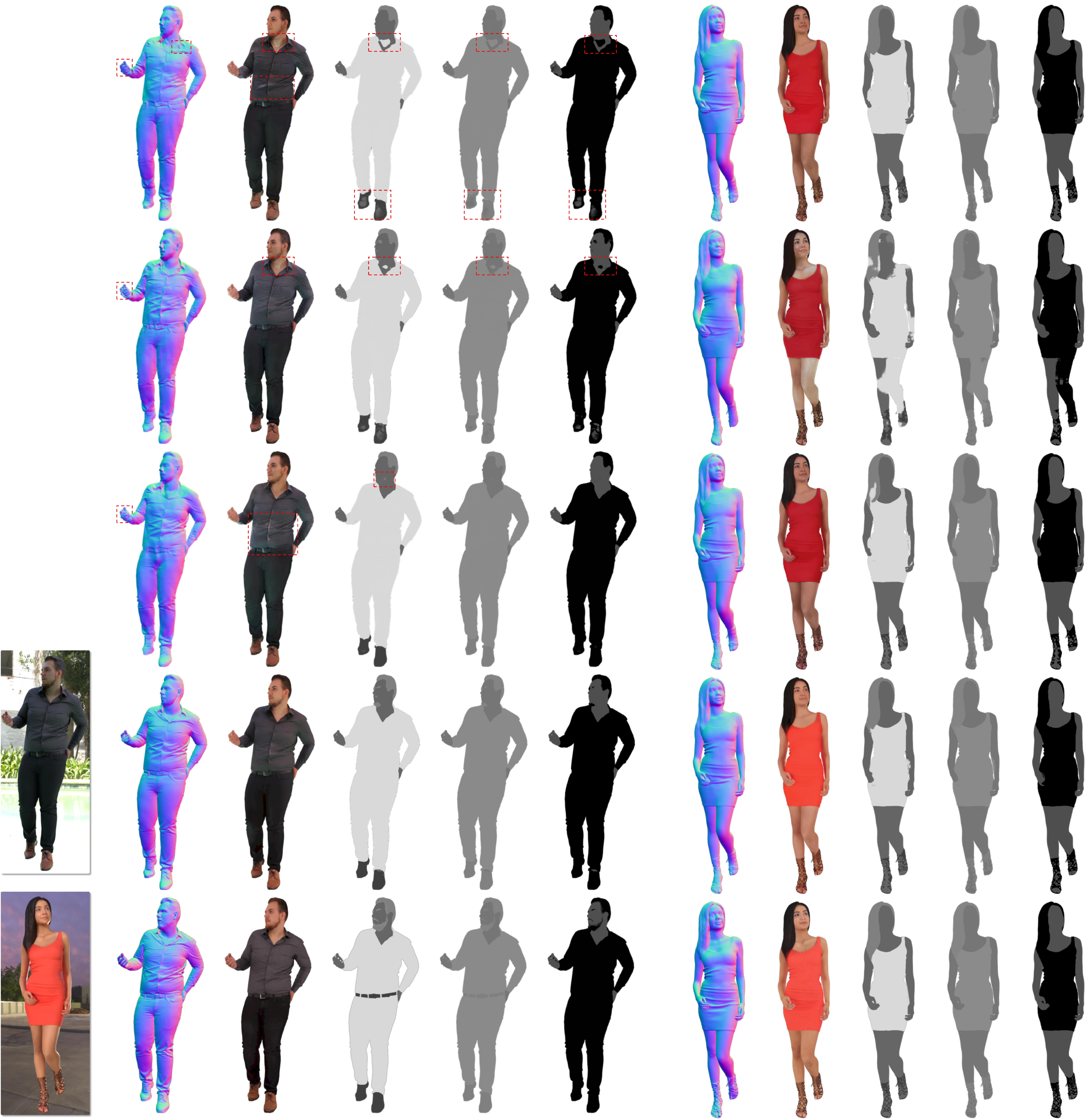}
  \begin{picture}(0,0)
  	\tiny
  	\put(-94,230){\rotatebox{90}{\makebox(0,0)[c]{w/o Guid}}}
  	\put(-94,180){\rotatebox{90}{\makebox(0,0)[c]{w/o P-Guid}}}
  	\put(-94,130){\rotatebox{90}{\makebox(0,0)[c]{w/o \FusionFullShort}}}
  	\put(-94,90){\rotatebox{90}{\makebox(0,0)[c]{Full}}}
  	\put(-94,40){\rotatebox{90}{\makebox(0,0)[c]{GT}}}
  \end{picture}
  \begin{picture}(0,0)
  	\tiny
  	\put(25,230){\rotatebox{90}{\makebox(0,0)[c]{w/o \Dino}}}
  	\put(25,180){\rotatebox{90}{\makebox(0,0)[c]{w/o \HuMat}}}
  	\put(25,132){\rotatebox{90}{\makebox(0,0)[c]{w/o \AFHR}}}
  	\put(25,87){\rotatebox{90}{\makebox(0,0)[c]{Full}}}
  	\put(25,40){\rotatebox{90}{\makebox(0,0)[c]{GT}}}
  \end{picture}
  \begin{flushleft}
  	\small
  	\vspace{-1.em}
    \hspace{0.8em}(a)
    \hspace{1.0em}(b)
    \hspace{1.0em}(c)
    \hspace{1.0em}(d)
    \hspace{1.0em}(e)
    \hspace{1.0em}(f)
    \hspace{1.8em}(b)
    \hspace{0.5em}(c)
    \hspace{0.5em}(d)
    \hspace{0.5em}(e)
    \hspace{0.5em}(f)
  \end{flushleft}
  \end{minipage}
  \caption{
  	Ablation study on \HumatDataset dataset. 
  	(a) Input. (b) Normal. (c) Diffuse albedo. (d) Roughness. (e) Specular albedo. (f) Subsurface scattering.
  	}
 \label{img-ablation_on_syn_dataset}
\end{figure}
Notably, the improvement in material estimation also translates into better relighting performance. Since relighting quality directly depends on the physical accuracy of the recovered materials, the superior relighting scores further verify that \Ours produces more physically plausible material estimates rather than merely fitting individual map-wise metrics.

\textbf{Visual Comparison.}
As shown in Fig.~\ref{img-material_comparision_on_syn}, Fig.~\ref{img-material_comparision_on_real_3}, Fig.~\ref{img-relighting_comparision}, and Fig.~\ref{img-more_relighting_comparision}, \Ours produces more accurate and visually consistent material maps than previous methods. In particular, our method preserves finer local details, such as facial structure, clothing boundaries, and small-scale material variations, leading to sharper and more faithful normal, diffuse, and reflectance-related predictions.

Compared with prior methods, our results also exhibit a better decoupling of material and illumination. Specifically, the estimated diffuse albedo contains fewer residual shading artifacts, while the roughness, specular albedo, and subsurface scattering maps better reflect intrinsic material properties rather than being contaminated by observed lighting effects. This indicates that the proposed hybrid guidance maps and adaptive fusion mechanism help the network more effectively separate appearance, geometry, material, and illumination cues from a single input image.

Such improvements further translate into more realistic relighting results. In particular, our relit images show more plausible shading transitions and more accurate specular reflections, with clearer and more convincing highlights on skin. As a result, the rendered appearance is visually closer to the ground truth and exhibits stronger realism under novel lighting.

\subsection{Ablation Study}
\label{ablation_study}
To validate the effectiveness of the proposed hybrid guidance and adaptive fusion design, we conduct six groups of ablation experiments, as shown in Table~\ref{table-ablation_on_syn} and Fig.~\ref{img-ablation_on_syn_dataset}. Specifically, we evaluate: (1) removing the guidance branch (``w/o Guid"), (2) using only the input image as guidance without geometric, structural, or prior-material maps (``w/o P-Guid"), (3) removing \FusionFullShort ~and replacing it with direct feature concatenation (``w/o \FusionFullShort"), and (4) removing the pre-trained priors from \Dino, \HuMat, or \AFHR.

\textbf{Effect of the guidance branch.}
Comparing ``Full" with ``w/o Guid" in Fig.~\ref{img-ablation_on_syn_dataset} and Table~\ref{table-ablation_on_syn} shows that removing the guidance branch degrades all material maps. This verifies that explicitly injecting external guidance helps reduce the ambiguity of single-image human material estimation. In particular, SMPL-derived cues improve normal estimation by providing more complete facial and body structure, while Densepose further benefits roughness, specular albedo, and subsurface scattering through body-part-aware semantic constraints.

\textbf{Effect of hybrid prior guidance.}
The setting ``w/o P-Guid" retains the guidance branch but uses only the input image as guidance. Its performance is better than ``w/o Guid" but consistently worse than ``Full", indicating that the branch itself is useful, while the full set of geometric, structural, and prior-material guidance provides additional physical and semantic constraints for more accurate material estimation.

\textbf{Effect of \FusionFullShort.}
Without \FusionFullShort, we directly concatenate the encoded input and guidance features and fuse them only at the deep layer. Compared with this variant, ``Full" achieves better results, showing that simple concatenation is insufficient for exploiting heterogeneous guidance cues. In contrast, MAFFM adaptively emphasizes texture-dominant and semantic dominant guidance at appropriate decoding stages, leading to more effective feature integration.

\textbf{Effect of pre-trained priors.}
Removing \Dino causes a noticeable overall performance drop, confirming that its latent visual features improve image representation for material decoding. Removing either \HuMat or \AFHR also degrades the results. This demonstrates that prior material predictions from pre-trained models provide effective guidance, while \HuMat and \AFHR remain complementary due to their different data sources and prior characteristics.

\subsection{Limitations}
\begin{figure}[!h]
	\centering % 居中
	\begin{minipage}[b]{\linewidth} % 整体划分一个页面
		\centering % 居中
		\newcommand{\myvspace}{1.0 pt} % 定义行之间的空隙
		\newcommand{\widthOfFullPage}{1} % 1 / 列数
		\newcommand{\widthOfMiniPage}{0.84}
		\newcommand{\format}{png}
	\includegraphics[width=\widthOfMiniPage\linewidth]{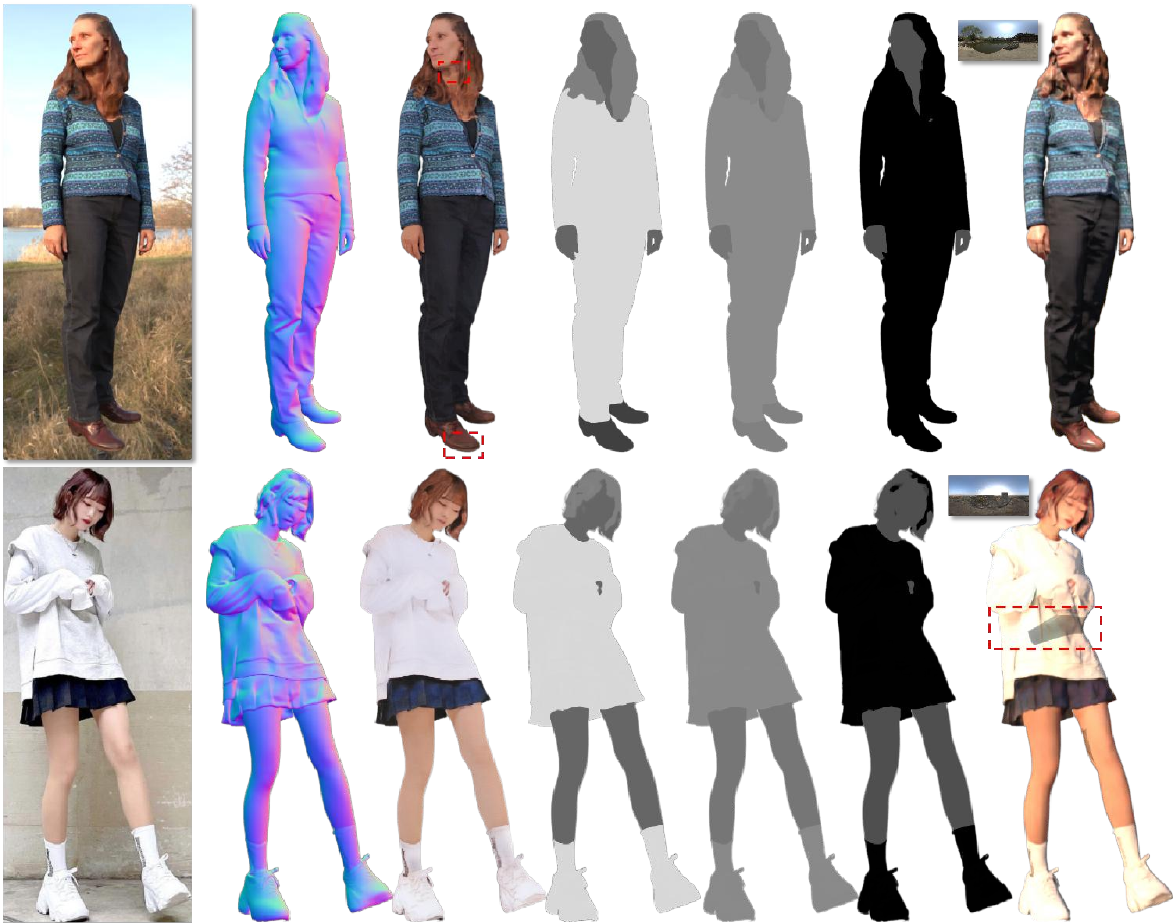}
	\begin{flushleft}
		\small
		\vspace{-0.5em}
		\hspace{2.8em}Input
		\hspace{2.2em}N
		\hspace{2.1em}D
		\hspace{1.9em}R
		\hspace{1.9em}S
		\hspace{1.4em}SSS
		\hspace{0.7em}Relit
	\end{flushleft}
	\end{minipage}
	\caption{
		Limitation. ``N'', ``D'', ``R'', ``S'', ``SSS'', and ``Relit'' denote normal, diffuse albedo, roughness, specular albedo, subsurface scattering, and relighting results, respectively.
	}
	\label{img-limitation}
\end{figure}

When the input image contains strong reflections or shadows, the estimated diffuse albedo may suffer from highlight and shadow artifacts. We consider a robust delighting preprocessing as a feasible solution.
Self-occlusion shadows can be calculated via ray tracing based on SMPL (``Relit" in Fig.~\ref{img-limitation}) . However, due to the gap between SMPL and the real geometry, these calculated shadows are biased. We therefore do not adopt them in the above rendering results.

\section{Conclusion}
In this work, we presented \Ours, a hybrid-prior-guided framework for single-image human PBR material estimation. Our method addresses the severe ambiguity of this task by introducing guidance maps that encode complementary cues from appearance, geometry, body structure, and prior material predictions. To better exploit the heterogeneous nature of these cues, we further proposed \FusionFullLow~(\FusionFullShort), which adaptively fuses guidance and decoder features at different stages. Extensive experiments on both \HumatDataset and real data demonstrate that our method achieves superior performance in material estimation and produces more realistic relighting results.

\vspace{1.5em}
\noindent\textbf{Acknowledgments}

This work is partially supported by the National Natural Science Foundation of China (No. 62372336), the Key Research and Development Program for Technological Innovation Project of Hubei Province (No. 2025B-AB020) and the Fundamental Research Funds for the Central Universities (2042026kf0016).

% Note: If you are using BibTeX, please use the following code:
\bibliographystyle{spmpsci}
\bibliography{mybib} 
% where ''bib_CGIconf'' has to be replaced by the name of your BibTeX
% file (without the .bib extension). 

\end{document}